# An AI-Guided Data Centric Strategy to Detect and Mitigate Biases in Healthcare Datasets

**Authors:** Faris F. Gulamali[1], Ashwin S. Sawant[1,2], Lora Liharska[1], Carol R. Horowitz[3], Lili Chan[1,2], Patricia H. Kovatch[4], Ira Hofer[1,2,5], Karandeep Singh[6], Lynne D. Richardson[3], Emmanuel Mensah[7], Alexander W Charney[1], David L. Reich[5], Jianying Hu[2,8], Girish N. Nadkarni[1,2]*

**Affiliations:**

1. The Charles Bronfman Institute of Personalized Medicine, Icahn School of Medicine at Mount Sinai; New York, 10029, USA.

2. The Division of Data Driven and Digital Medicine, The Samuel L Bronfman Department of Medicine, Icahn School of Medicine at Mount Sinai; New York, 10029, USA.

3. Institute for Health Equity Research, Icahn School of Medicine at Mount Sinai; New York, 10029, USA.

4. Department of Genetics and Genomic Sciences, Icahn School of Medicine at Mount Sinai; New York, 10029, USA.

5. Department of Anesthesiology, Perioperative and Pain Medicine, Icahn School of Medicine at Mount Sinai; New York, 10029, USA.

6. Department of Learning Health Sciences, University of Michigan Medical School, Ann Arbor, MI

7. Department of Medicine, Beth Israel Deaconess Medical Center / Harvard Medical School; Boston, 02215, USA.

8. IBM Research, Yorktown Heights, NY

*Corresponding author:

Girish N Nadkarni, MD, MPH

girish.nadkarni@mountsinai.org

Word count (main text): 2,976

Word count (abstract): 350

## Abstract

The adoption of diagnosis and prognostic algorithms in healthcare has led to concerns about the perpetuation of bias against disadvantaged groups of individuals. Deep learning methods to detect and mitigate bias have revolved around modifying models, optimization strategies, and threshold calibration with varying levels of success. Here, we generate a data-centric, model-agnostic, task-agnostic approach to evaluate dataset bias by investigating the relationship between how easily different groups are learned at small sample sizes (AEquity). We then apply a systematic analysis of AEq values across subpopulations to identify and mitigate manifestations of racial bias in two known cases in healthcare – Chest X-rays diagnosis with deep convolutional neural networks and healthcare utilization prediction with multivariate logistic regression. AEq is a novel and broadly applicable metric that can be applied to advance equity by diagnosing and remediating bias in healthcare datasets.

# Main

The adoption of diagnosis and prognostic algorithms in healthcare has led to concerns about the perpetuation of bias against disadvantaged groups of individuals. For example, an algorithm used to predict healthcare costs exhibits bias against Black patients [1]. Such bias arises from the longstanding societal and structural biases and disparities, reflected in real-world data used to develop algorithms [2]. These biases exist even in complex algorithms trained using diverse populations. For example, a standard computer vision model applied to chest radiographs resulted in selective underdiagnosis of pneumonia in people who are Black, Hispanic/Latino, or recipients of Medicaid [3].

Algorithms developed using data from systems with longstanding discrimination and inequities tend to recapitulate those biases [4–8]. The primary focus of mitigating model bias has been during training, optimization or post-processing [9–11]. However, in many healthcare machine learning tasks, data curation and collection remain essential depending on the depth of phenotyping and number of samples [12–14]. There is increasing recognition of the need to address bias earlier in the algorithm development pipeline, and more specifically, at the data collection and curation step [15–17].

In this paper, we incorporate two separate but related definitions of bias–predictive bias and residual unfairness. When the task involves identifying a diagnostic finding such as pneumonia on a Chest X-ray, we define predictive bias as differences in performance against, or in favor of, a subpopulation [3,18]. However, if the underlying dataset is biased, models can exhibit residual unfairness even after fairness-metric based optimization [19–22]. Algorithms themselves can create bias when trained with "complete" data that contains minoritized populations, simply because these groups may require different models. For example, black patients utilize healthcare less than white patients, and consequently, a model trained on both black and white patients that stratifies risk of healthcare utilization characterizes black patients with a large number of significant comorbidities as requiring fewer resources, even after fairness-based optimization strategies [1].

We provide evidence that machine learning models can both reduce predictive bias and residual unfairness with AEquity, an AI-guided data curation strategy that we introduce to evaluate dataset bias by functioning as a stratified sample size determination tool that can be applied directly to a small subset of the data (approximately 3%) or, in more complex cases, to the latent space of a deep learning model trained on a small subset of the data (approximately 3%). Moreover, we provide an easy-to-use implementation in the form of a Python library (**Supplementary Methods**).

Generally, the task can be described as the following: assume we have a small subset of collected data containing two distinct groups with features X and potential outcomes y. The first task is to determine which outcome should be measured to minimize residual unfairness towards a protected group. The next task is to determine which subset of data needs to be collected to reduce differences in performances against, or in favor of, a group based on a protected characteristic.

## A Framework for Understanding Dataset Bias

In the following manuscript, we use "groups" to refer to various subpopulations, for example race, gender, or insurance status. We will use "outcome" to refer to diagnostic findings like "pneumonia" and "edema".

Following the categories defined in [2], we characterize dataset biases into main types: sampling bias; complexity bias; and label bias ([Figure 1](Figure 1)). Sampling bias arises when there is non-random sampling of patients, and this results in a lower or higher probability of sampling one group over another. Under-representation can negatively impact generalization performance. An example of this is under-representation of Black patients with a diagnosis of pneumothorax. In complexity bias, a group presents more heterogeneously with a diagnosis, and consequently, one group exhibits a class label with greater complexity than another group. Subsequently, generalization performance on that group is worse than other groups for the same class. Third, label bias occurs when labels are placed incorrectly at different rates for different groups and can lead to increased misclassification errors for the affected group. Black patients are more frequently labeled as requiring fewer healthcare resources despite having the same number of comorbidities because of lack of access to care.

In this manuscript, we describe a deep learning-based metric that can be used for the characterization and mitigation of these three biases (sampling, complexity and label) in healthcare datasets. AEquity works by appending a simple compressive network to a dataset or the latent space of existing deep learning models to generate a single value characterizing a label with respect to a group and the priors established by the model. In general, a higher AEq value for a group within a given diagnostic label with respect to a machine learning model suggests that generalization is more difficult and more samples are needed for learning [23]. For example, AEq is a metric that is calculated at the intersection of a diagnostic label like "pneumonia", a group label like "Black", and a machine learning model like "ResNet-50." AEq is reported as $\log_2$(sample size estimate). AEq values are only meaningful when constrained to comparing groups with the same label in the dataset, and therefore interpretation of the metric is invariant to architecture of the deep learning model, the compressive network or the dataset type or characteristics [24–26].

AEquity-based data curation was achieved through a multi-step pipeline (Figure 2). First, a group-balanced subset of the patient data is collected and this data is labeled. Next, AEquity values by group for each outcome are calculated. Finally, AEquity is calculated with respect to each outcome when groups are combined into a joint dataset ([Figure 2](#)). When predictive bias is driven primarily by sampling bias in the dataset, combining groups drives the AEq value at or below the over-sampled data (**Supplementary Methods**). In sampling bias, group balancing is sufficient to mitigate the bias because the data is equitably represented within the model. When complexity bias was the only type of dataset bias, combining the groups resulted either in an increase in the value of AEq or an AEq closer to value that had been higher prior to the combination. In complexity bias, prioritizing data collection of the protected population, due to its relative heterogeneity compared to the over-represented group, is necessary to mitigate bias. If the model exhibits residual unfairness, then the AEq will be different for two groups that exhibit the same label. Selecting a better label based on domain knowledge can mitigate the label bias.

# Detection and Mitigation of Racial Bias via Outcome Selection in Electronic Healthcare Record Datasets at Small Sample Sizes.

We examined a publicly available tabularized EHR dataset of 49,618 primary care patients enrolled in risk-based contracts from 2013 to 2015, from an unspecified large academic hospital, described in Obermeyer et al[1]. Data included demographics, visit information. In that paper, the authors examined an algorithm which predicted total healthcare costs for a patient as a proxy for their predicted healthcare needs. They categorized individuals at or above the 50th percentile of predicted outcome as 'high risk' and these patients would be screened for a care management program, whereas patients at the 95th percentile would be automatically enrolled. They found that Black patients screened for the high-risk category were significantly sicker compared to White patients in that category. This is a form of bias because it resulted in a reduced allocation of resources to Black patients in comparison with their healthcare needs. They found that this discrepancy persisted when patients were categorized based on another predicted cost-based metric-avoidable costs, and despite calibration to various fairness metrics. However, the discrepancy disappeared if predicted active chronic conditions were used for categorization.

AEquity-based data curation was achieved through a multi-step pipeline. First, a race-balanced subset of the patient data is collected (n=1,024, 3% of the total available data) and this data is labeled according to each of the available labels (active chronic conditions, avoidable costs, and total costs). We calculated AEquity by race (Black, White), for each risk category (high risk, low risk), for each of the three outcomes (total costs, avoidable costs, active chronic conditions). (Table 1). To determine if Black and White patients in each risk category were similar, we then calculated the difference in AEquity between races, for each risk category, for each outcome.

First, the difference in AEquity was statistically significant across outcome measures for the population designated as high-risk (ANOVA, P = $1.59 \times 10^{-8}$). When predicted cost-based metrics (total costs, avoidable costs) were used to designate patients at highest risk, there was a significant difference in the AEquity metric between Black and White patients ([Figure 3A, 3B](); [Table 1]()). For example, with total costs, the difference was 0.47 (95% CI, 0.36-0.52). Thus, Black patients predicted to be at high risk were significantly different from their White counterparts which led to inequitable resource allocation.

However, the difference between Black and White patients categorized to be at high risk disappeared when using predicted active chronic conditions vs. predicted costs as the outcome measure--the difference in AEquity was not statistically significantly different from 0 [0.06 (95% CI, -0.04 to 0.16)], indicating that Black and White patients in that risk category were similar. Thus, the guided choice of a better outcome measure to identify patients at high risk mitigated bias.

We conducted additional analyses in patients at lower risk and found that the differences in AEquity across races were all statistically significant ([Figure 3B]()). In the low risk group, black patients display more heterogeneous underlying characteristics than white patients. We also observed that the differences in the AEquity metric were smaller when using active chronic conditions than avoidable costs or total costs, indicating mitigation of bias for this lower-risk group despite the increased heterogeneity ([Table 1]()).

# Detection at Small Sample Sizes and Mitigation of Under-diagnosis Bias in Chest X-ray

We next evaluated AEquity on MIMIC-CXR, a publicly available dataset containing 377,110 chest radiographs of 65,379 patients from the Beth Israel Deaconess Medical Center between 2011 and 2016 [27]. We also used two additional publicly available chest radiograph datasets to validate our findings – CheXpert containing 224,316 chest radiographs from 65,240 patients, and ChestX-ray14 containing 108,948 chest radiographs from 32,717 patients. MIMIC-CXR and CheXpert each contain 14 binary diagnostic findings, ChestX-ray14 contains 15, and all three contain an additional binary label for radiographs with no positive diagnostic finding. ([Supplementary Table 1](#)) [28,29]. We generated a subset from MIMIC-CXR by including only postero-anterior radiographs of patients who self-identified as either Black or White, and with either one of the nine most frequently encountered diagnostic findings (atelectasis, cardiomegaly, consolidation, edema, effusion, enlarged cardiomediastinum, opacity, pneumonia, pneumothorax) or 'no finding'. In Seyyed-Kalantari et al. [3], a deep convolutional neural network is naively trained on Chest X-rays belonging to the MIMIC-CXR dataset and evaluated on protected populations. The model demonstrates differences in performance on protected populations. Moreover, they find that methods which strive to achieve better worst-group performance do not outperform balanced empirical risk minimization [9,16,30,31].

AEquity-based data curation was achieved through a multi-step pipeline. First, a balanced subset of the patient data is collected (n=1,024, 3% of the total available data, 512 black patients, 512 white patients) and this data is labeled according to each of the available labels. We calculated AEquity by race (Black, White), for each diagnostic finding for a given model (AlexNet, RadImageNet, Vision Transformers). We use AEquity to guide data collection - either choosing to group-balance or prioritize data collection of a specific group to a total of 30,000 samples, a sample size sufficient to produce performance comparable state-of-the-art methods (Table S2). We split the subset 60%-20%-20% into training samples, validation samples, and test samples by patient, ensuring no patients overlapped, and trained the model with PyTorch lightning [32]. We show how AEquity can improve performance over naive data collection (Figure 4, Table 2), and benchmark against balanced empirical risk minimization (Table 3). We validate against threshold-independent metrics such as area-under-the-curve as well as fairness metrics such as sensitivity, specificity, FPR, FDR, FNR in intersectional populations (Table S3). We bootstrap each experiment 50 due to computational constraints [33–35].

First, we observe that a higher frequency of occurrence for a diagnosis corresponded to a lower AEq value across Black and White patients (Figure 4c). This indicates that labels with more samples, like lung opacity (AEq ≈ 7.5), generally trend towards lower AEqs, whereas labels with fewer samples such as consolidation (AEq ≈ 9.8), enlarged cardio-mediastinum (AEq ≈ 9.9) or pneumothorax (AEq ≈ 9.95) trend towards having higher AEqs. We posit that the latter group of diagnoses may have the highest potential for bias because of difficulty in generalization. Second, we noticed that six out of the nine diagnoses, accounting for 71% of the positive samples, demonstrate higher AEq values in Black patients than in White patients. Higher AEq values indicate that, in general, models built on chest X-rays from Black patients for Black patients generalize more poorly than models built on chest X-rays from White patients for the purpose of diagnosing White patients. In other words, more samples from Black patients than White patients are required to train a model that achieves equitable performance.

First, we examined pneumothorax, Chest X-rays from Black individuals diagnosed with pneumothorax have a significantly larger AEq values than chest X-rays belonging to White individuals with the same diagnoses ($AEq_{Black, Pneumothorax}$ = 10.10; 95% CI: (10.01, 10.18),

$AEq_{White, Pneumothorax}$ = 9.50; 95% CI: (9.44, 9.58); P < 0.05). For the diagnosis of pneumothorax, the AEq of the joint dataset is smaller than the AEq of chest X-rays belonging to each subgroup, which implies that ($AEq_{Joint, Pneumothorax}$ = 9.44, 95% CI: (9.35, 9.52)). Subsequently, we see that adding dataset diversity by equitably sampling from each race was associated with an improvement in classifier performance for Black individuals when compared to the naïve approach (Bias Reduction (Absolute) = 2.43 x $10^{-2}$; 95% CI: (2.22 x $10^{-2}$, 2.59 x $10^{-2}$, Bias Reduction (%) = 49.37%, 95% CI: (49.3, 49.5), P < 0.05). Second, we examined edema, where the joint AEq was higher than the AEq for either race, consistent with complexity bias. For edema, $AEq_{Joint}$ is larger than both $AEq_{Black}$ and $AEq_{White}$ ($AEq_{Joint, Edema}$ = 9.48; 95% CI: (9.37, 9.58), $AEq_{Black, Edema}$ = 8.78; 95% CI: (8.69, 8.86); $AEq_{White, Edema}$ = 9.12; 95% CI: (9.08, 9.20), P < 0.05). Prioritizing data collection from Black patients better captured the group-specific presentations, and this strategy was associated with a model that generalized better to Black patients when compared to a naïve approach (Bias Reduction (Absolute) = 1.95 x $10^{-2}$; 95% CI: (1.90 x $10^{-2}$, 2.01 x $10^{-2}$, Bias Reduction (%) = 60.02%, 95% CI: (59.98, 60.06), P < 0.05).The bias for each diagnostic finding in the chest radiograph dataset decreased by between 29% and 96.5% following intervention (Table 1; Figure 2D).

Finally, we find that this 1) AEquity remains valid for sex, 2) AIquity-guided data curation is effective for individuals at the intersection of more than one disadvantage group and 3) Targeted data collection and curation is robust to choice of fairness metric. When evaluating sex, the joint AEq score for all 9 diagnostic findings was not significantly different from, or lower than the AEq score for females, indicating that predictive biases are primarily driven by sampling bias ([Figure S1](#)). Larrazabal et al [16] has previously demonstrated that group balancing is the optimal approach to mitigate gender biases for models trained on MIMIC Chest X-rays. Second, demonstrate the validity of AEq on intersectional populations, who are at greater risk of being affected by systemic biases [36,37], and we show that this technique is robust to the choice of fairness metrics. When we examined Black patients on Medicaid, at the intersection of race and socioeconomic status, we find that AEquity-based interventions reduced overall false negative rate in false negative rate by 33.3% (Bias Reduction Absolute = 1.88 x $10^{-1}$; 95% CI (1.4x$10^{-1}$, 2.5x$10^{-1}$); Bias Reduction (%) 33.3% (95% CI, 26.6-40.0)), Precision Bias by 7.50x$10^{-2}$; 95% CI (7.48x$10^{-2}$, 7.51x$10^{-2}$); Bias Reduction (%) 94.6% (95% CI, 94.5-94.7%); False Discovery Rate by 94.5% (Absolute Bias Reduction = 3.50x$10^{-2}$; 95% CI: (3.49x$10^{-2}$, 3.50x$10^{-2}$) ([Figure S3](Figure S3)).

## Discussion

Bias in healthcare algorithms is increasingly becoming a focus for regulators. For example, the Good Machine Learning Practice guidance from the FDA and EMA emphasizes the importance of ensuring that datasets are representative of the intended patient population [38]. However, this alone is not sufficient because significant bias can arise even when standard machine learning methods are applied to diverse datasets. Integration of the AEquity metric into the algorithm development pipeline may help clarify the cause of this bias and mitigate it.

The impetus of equity in machine learning is increasingly being placed on model developers. The US Department of Health and Human Services has proposed a rule under Section 1557 of the Affordable Care Act to ensure that "a covered entity must not discriminate against any individual on the basis of race, color, national origin, sex, age, or disability through the use of clinical algorithms in decision-making". More recently, executive order placed by President Biden directly placed the liability of model equity on developers and deployers [39]. Adherence to ML/AI guidelines as well as to this executive order remains an important task and methods to enforce regulations and data audits is still underway. Therefore, the need to develop tools to identify and remediate potential bias in healthcare algorithms is urgent.

The model development pipeline involves task identification, data collection, data curation, model training, model validation, and model deployment, each of which plays an important role in whether a machine learning model adds value to the healthcare system or propagates the very inequities that it seeks to address [40–43]. The ability of AEquity to quantify and address bias at small sample sizes makes it an attractive addition to the toolset, akin to sample size determination but for bias [10]. There remains many other pathways in which bias can impact model development, and how practitioners, from start to finish, can address these biases [44].

Data-guided strategies and AEquity face strong challenges due to the lack of representation in many datasets [45,46]. Moreover, once a gap has been identified, targeted data collection of protected populations raises ethical concerns given the long history of mistrust [4–8,47]. Nevertheless, acknowledging the existence of these biases through quantitative measures such as AEquity can build trust, a crucial step towards mitigating some of these biases at a grassroots level [48,49].

While our results are promising, there are some limitations of AEq and directions for future inquiry. In our study we primarily focus on two case examples - underdiagnosis and under-allocation of resources. We focus on these examples because under-diagnosis can lead to under-allocation of resources and vice versa, a vicious cycle of healthcare inequity. However, utilization of AEquity on a broader scale across all phenotypes, protected groups, data modalities and model types is essential to further validating our approach.

Third, AEq in its current form focuses primarily on classification tasks with an unregularized latent space. A growing number of algorithms, however, are: a) using various forms of regularization and attention to produce more informative latent spaces [50,51]; and b) being used for generation and representation learning rather than simply classification or regression [52] and c) are multi-modal [53]. Extending AEq to these latent spaces and generative models may help investigate and mitigate bias in those settings.

Fourth, the labels for the MIMIC-CXR dataset were obtained using the CheXpert system of natural language processing of radiology reports, which is known to have a substantial and varying amount of mislabeling for different diagnoses [29]. Results from Ghassemi et al suggest there is no statistically detectable label bias by race in MIMIC-CXR for the 'no finding' label [9]. We cannot exclude the possibility of label bias related to the rate of mislabeling being different for different groups for other diagnoses. The application of frameworks for dataset development which include expert labeling of the most informative samples may help mitigate label bias [54].

In summary, we present AEq, a novel, deep learning based metric that may be valuable for disentangling and quantifying various manifestations of bias at the dataset level. We show how it can be used to suggest specific, effective mitigation measures. Finally, we demonstrate its robustness by applying it to different datasets and models, intersectional analyses and measuring its effectiveness with respect to a range of traditional fairness metrics.

# Acknowledgements

The authors would like to acknowledge Jill Gregory for support in illustrating the figures and the staff at the Minerva High Performance Computing Cluster for maintaining the servers used to run the experiments.

**Funding:** The National Center for Advancing Translational Sciences and the National Institutes of Health had no role in study design, data collection or analysis, preparation of the manuscript, or the decision to submit the manuscript for publication.

Data and materials availability:

The publicly available images can be access via the PyTorch library ([torchvision — Torchvision main documentation (pytorch.org)](#)). The MIMIC-CXR version 2.0 dataset is publicly available by registration at https://physionet.org/content/mimic-cxr/2.0.0/. The CheXpert dataset is publicly available by registration at https://stanfordmlgroup.github.io/competitions/chexpert/. The NIH-CXR dataset is publicly available at https://nihcc.app.box.com/v/ChestXray-NIHCC. The dataset used in the Obermeiyer analysis is available in the supplement ([Dissecting racial bias in an algorithm used to manage the health of populations | Science](#)). The code for the analyses are available here ([Nadkarni-Lab/AEquitas: Deep Learning Based Metric to Mitigate Dataset Bias (github.com)](#), and the instructions for usage are included in the supplement

# Figures

Figure 1

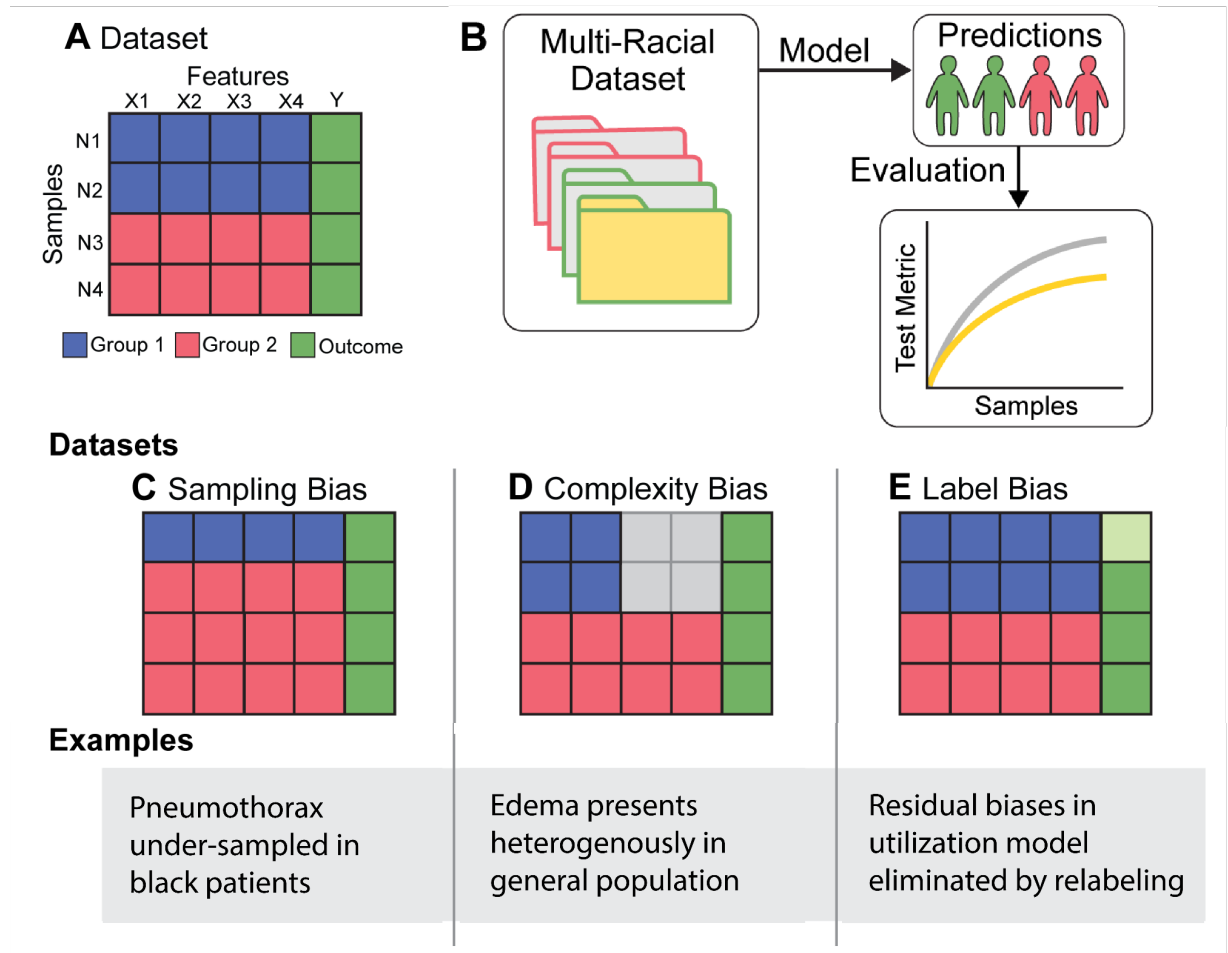

**Figure 1. Description of Different Biases and Proposed Role of AEquity in Identifying and Mitigating Bias.** A. A prototypical dataset containing samples from two groups each of which has four informative features. B. When trained in a similar manner on a more diverse dataset, the resulting model is biased because it performs worse for a particular group. C. Sampling bias occurs when one group is dramatically under-represented compared to a second group. D. Complexity bias results when one group presents more heterogeneously than a second group; for example patients with edema on chest radiographs present more heterogeneously in the general

population than in each subgroup. E. Label bias occurs when mislabeling burdens one group more than another; Light green represents a higher burden of mislabeling.

## Figure 2

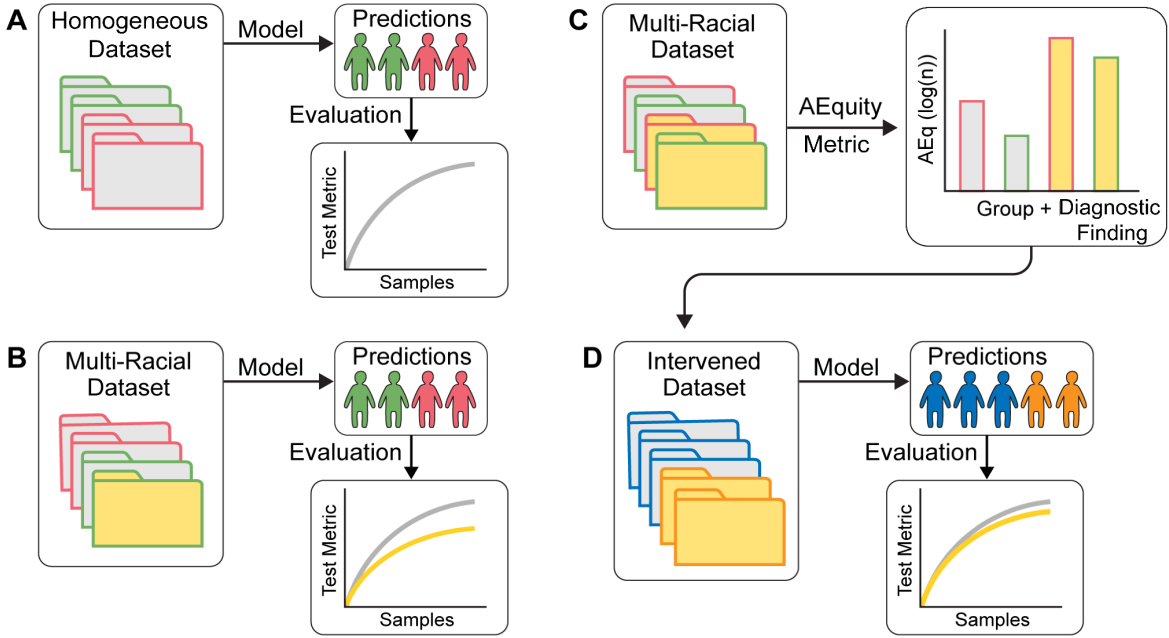

**Figure 2. Application of AEquity to detect and mitigate diagnostic biases in MIMIC-CXR.**
A. Red – White individuals; Blue – Black individuals. B. Frequency of diagnostic findings by race. Purple – average of White and Black. C. AEquity values (log(N)) by diagnosis and race. D. Effect of data-centric mitigation intervention on bias against Black patients. E. Decrease in bias as measured by different fairness metrics in Black patients on Medicaid.

# Figure 3

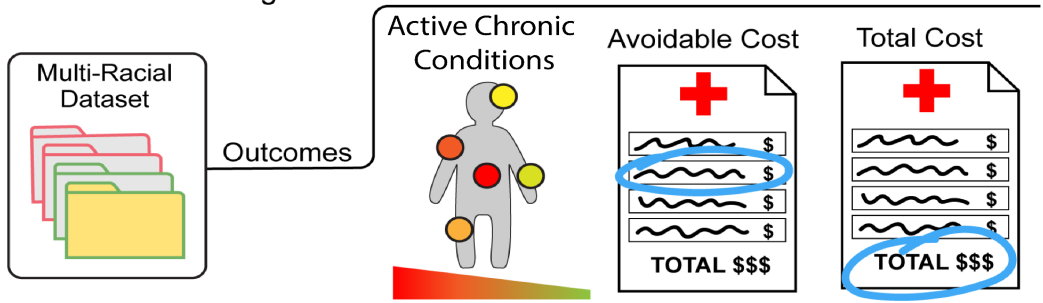

**A** Cost Utilization Algorithm

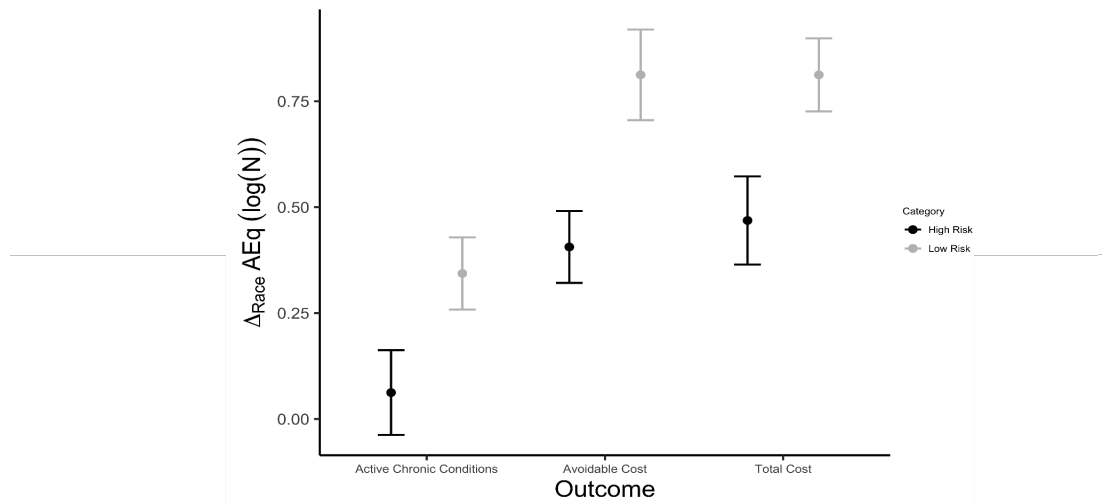

**B** AEquity Calculations

**Figure 3. Detecting bias in prediction of healthcare needs with the AEquity metric.**
A. A dataset with claims data used to calculate a score based on number of active chronic conditions, and two cost-based metrics. B. Difference in AEquity values between Black and White patients for comorbidity- and cost-based outcomes, stratified by risk level.

Figure 4

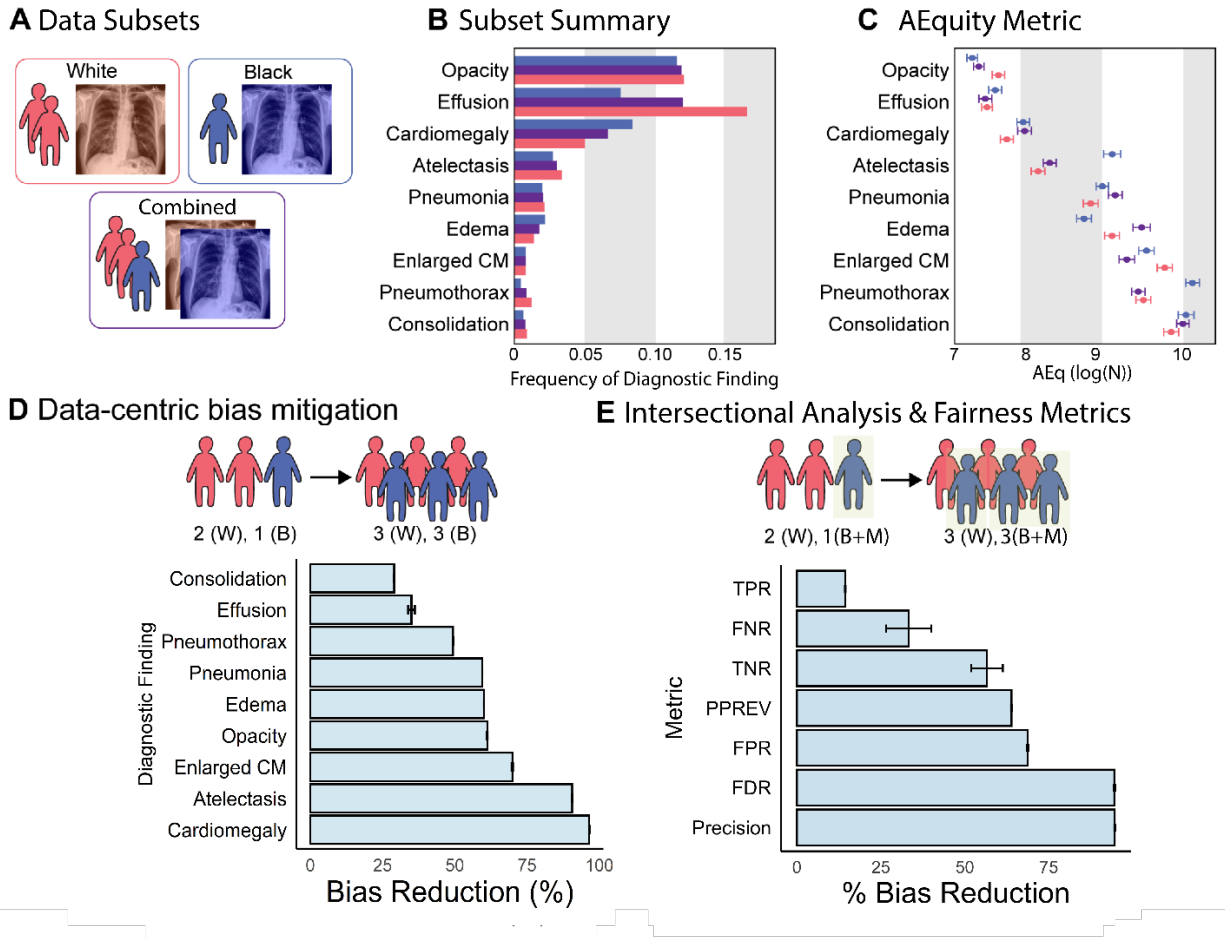

**Figure 4. Application of AEquity to detect and mitigate diagnostic biases in MIMIC-CXR.**
A. Red – White individuals; Blue – Black individuals. B. Frequency of diagnostic findings by race. Purple – average of White and Black. C. AEquity values (log(N)) by diagnosis and race. D. Effect of data-centric mitigation intervention on bias against Black patients. E. Decrease in bias as measured by different fairness metrics in Black patients on Medicaid.

# Tables

## Table 1

**Table 1**. Measurement of the discrepancy between Black and White patients for each risk score in the EHR dataset using the AEquity metric.

| Risk Metric | Risk Score | AEquity Metric (95% Confidence Interval) |
|---|---|---|

| Avoidable Costs | Low Risk | 0.81 (0.71-0.92) |
| Avoidable Costs | High Risk | 0.41 (0.32-0.49) |
| Total Cost | Low Risk | 0.81 (0.73-0.90) |
| Total Cost | High Risk | 0.47 (0.36-0.57) |
| Active Chronic Conditions | Low Risk | 0.34 (0.26-0.43) |
| Active Chronic Conditions | High Risk | 0.06 (-0.04 to 0.16) |

We show the AEquity metric calculated with respect to each risk metric in both the predicted low risk and high-risk population. When stratifying via avoidable costs and total costs, the AEquity metric is elevated, but when stratifying by active chronic conditions, the AEquity metric is not elevated indicating that active chronic conditions is a better choice of outcome to mitigate bias.

Table 2

**Table 2.** Measurement of bias and intervention effect by diagnosis in the radiograph dataset with RadImageNet (ResNet-50 with pre-training on Medical Images)

| Diagnosis | Naive AUROC in White Patients (95% CI) | Naive AUROC in Black Patients (95% CI) | Post-Intervention AUROC in White Patients (95% CI) | Post-Intervention AUROC in Black Patients (95% CI) | Bias against Black patients (Pre-Intervention, %) | Bias against Black patients (Post-Intervention, %) | Bias Reduction (%) |
|---|---|---|---|---|---|---|---|
| Pneumothorax* | 0.73 (0.72,0.74) | 0.68 (0.67, 0.69) | 0.73 (0.72, 0.74) | 0.70 (0.68,0.72) | 7.5 | 3.8 | 49.4 (49.3, 49.5) |
| Atelectasis* | 0.69 (0.68, 0.69) | 0.64 (0.64, 0.65) | 0.70 (0.70, 0.70) | 0.69 (0.69,0.69) | 6.4 | 0.6 | 90.6 (90.5, 90.7) |
| Edema* | 0.75 (0.75,0.75) | 0.72 (0.71, 0.72) | 0.78 (0.78, 0.78) | 0.79 (0.79,0.80) | 5.0 | 2.0 | 60.0 (59.9, 60.1) |
| Consolidation* | 0.67 (0.67,0.67) | 0.65 (0.64, 0.66) | 0.63 (0.63, 0.63) | 0.61 (0.60,0.63) | 3.8 | 2.8 | 29.0 (28.8,29.2) |
| Pneumonia* | 0.67 (0.66,0.67) | 0.64 (0.64, 0.65) | 0.65 (0.65, 0.65) | 0.64 (0.64,0.65) | 3.8 | 1.5 | 59.5 (59.4, 59.6) |
| Enlarged CM* | 0.62 (0.61,0.63) | 0.60 (0.59, 0.60) | 0.61 (0.61, 0.62) | 0.61 (0.60,0.61 | 3.6 | 1.1 | 69.9 (69.6, 70.1) |
| Cardiomegaly* | 0.76 (0.76,0.76) | 0.75 (0.75, 0.75) | 0.76 (0.76, 0.76) | 0.76 (0.76,0.76) | 1.9 | 0.1 | 96.5 (96.4,96.6) |
| Pleural Effusion | 0.64 (0.64, 0.64) | 0.64 (0.64, 0.64) | 0.65 (0.65, 0.65) | 0.65 (0.65,0.65) | 0.3 | 0.2 | 35.0 (33.7,36.2) |
| Opacity | 0.69 (0.69, 0.69) | 0.71 (0.70, 0.71) | 0.69 (0.69, 0.69) | 0.70 (0.69,0.70) | -2.0 | -0.8 | 61.2 (61.0,61.4) |

We show AUROC before and after intervention with 95% Confidence Intervals, for Black and White patients. Bias is the difference in the AUROC between Black and White patients. With a dataset intervention, we reduce the discrepancy between the two groups and calculate percentage change in the discrepancy. Percentages have been rounded to one decimal place, and all other numbers have been rounded to two decimal places. * - diagnostic findings with bias against Black patients.

Table 3

| Label | Balanced ERM (%) | AEquity (%) | % Improvement over sub-optimal policy |
|---|---|---|---|
| Atelectasis | 0.85 (0.76, 0.94) | **0.51** (0.42, 0.60) | 40.07 (39.09, 41.05) |
| Pneumonia | 1.70 (1.64, 1.76) | **1.39** (1.33, 1.45) | 24.3 (24.26, 24.34) |
| Pneumothorax | 7.11 (7.00, 7.22) | **6.42** (6.31, 6.53) | 24.79 (24.61, 24.79) |
| Consolidation | **2.36** (2.24, 2.48) | **2.36** (2.24, 2.48) | 0.00 (-0.7, 0.7) |
| Edema | 2.11 (2.06, 2.16) | **2.03** **(1.98, 2.08)** | 3.77 (3.73, 3.81) |
| Effusion | 0.41 (0.35, 0.46) | **0.35** (0.30, 0.41) | 12.9 (10.61, 15.21) |
| Cardiomegaly | **0.18** **(-0.13, 0.49)** | 0.18 (-0.13, 0.49) | 0.66 (0.54, 0.78) |
| Enlarged CM | 3.59 (3.42, 3.76) | **3.12** (2.95, 3.29) | 13.1 (12.83, 13.35) |
| Opacity | 0.49 (0.38, 0.60) | **0.24** (0.13, 0.36) | 50.3 (50.10, 50.34) |

**Table 3.** Percentage Bias against black patients by method in a ResNet-50 naive to medical images.

# Supplement

## Materials and Methods

### Background on the AEquity metric

AEquity is rooted in information bottleneck theory. The core idea of information bottleneck theory is that a deep learning classifier mapping input X to output y does so by retaining the relevant information and eliminating the irrelevant information through a hidden layer h[1]. In contrast, an autoencoder reconstructs X, i.e., maps input X to input X through a hidden layer h and must retain all meaningful information. If y semantically encodes all the meaningful aspects of X, then the information loss between X and y through h in the classifier is minimal. The information, therefore I(X;y) is approximately equal to I(X;X) at all sample sizes, and rate of learning for the autoencoder and the generalizability of a deep learning model is similar[2]. Information flow is invariant to autoencoder architecture and loss, which makes this technique generalizable across different hyper-parameters[3].

We can subsequently utilize this approximation for sample size determination of deep learning models. A standard practice in traditional experimental design, sample size estimation takes place prior to data generation, and describes the minimum number of samples required to observe an effect of an independent variable on an outcome. The deep learning equivalent is the minimum convergence sample, which describes the minimum number of samples for an algorithm to start generalizing to unseen data. If $f_n$ is a deep learning algorithm that maps an input X to h to y trained on n samples where h is the latent representation, then the minimum convergence sample *n* is the number of training samples such that E[AUC ($f_n(X_{test}, y_{test})$)] > 0.5. In practice, this number cannot be directly observed without prospective validation because the test set, by definition, is not available until training is complete. We can, however, approximate the number of samples utilizing an autoencoder, which we define as $g_n$: X to h to X, where h is the latent representation. The minimum number of samples *n* such that E[$g_n(X_{train}, X_{train})$] < 1.0 is the minimum convergence sample estimate. Above the minimum convergence sample, our previous work has shown that the decrease in autoencoder loss is proportional to the gain in generalizability as measured by area-under-the-curve metric[4]. We have previously shown that the minimum convergence sample estimate remains a valid approximation of generalizability across model architecture, hyperparameter selections, dataset dimensions and dataset complexity.

Bias is marked by differences in generalization performance by group or class. Consequently, we examined the MCSE metric on subsets of the data consisting of individual groups and classes. The resulting *AEquity* metric is defined as

$$AEq_i := arg\ min_n \frac{d^2 L_{2_i}}{dn^2} \qquad \textbf{Equation 1}$$

where $L_{2_i}$ is the *i*[th] class/group-specific loss curve for a network trained on samples drawn from the entire dataset. A smaller *n* tends to indicate a better generalization performance, whereas a larger *n* indicates poor generalization performance when compared to other classes.

## Code Availability

The code for the analyses is available at https://github.com/Nadkarni-Lab/AEquity. The steps to use the repository are listed below.

1. Install requirements.

These requirements are publicly sourced such as

sklearn torch scipy torchvision pandas

pip install requirements.txt

2. Setup configuration in a config.yaml file.

```
data_path: ./data/custom_data.tsv # Contains path to data. Contains independent variables, demographics, and outcome variables.
demographics_col: demographics # Name of demographics variable in data_path.
outcome_cols: outcome_1  # Name of outcome variable in data_path
exclude_cols: None # name of columns to exclude if there are extraneous columns.
out_data: ./output/data.tsv # Output directory for AEq analyses.
bootstraps: 10 # Number of bootstraps. 30-50 is typically recommended for resolution at 5000 samples.
start_seed: 42 # Seed experiments.
input_dim: 149 # Number of independent columns in data_path
max_sample_size: 5000 # Max sample size to calculate from. Usually only require 128-512 samples.
root_dir: ./weights # Root directory to output weights, and other output files.
```

3. Run the measure and mitigate experiments.

```
python measure_disparity.py --config config.yaml
python mitigate_disparity.py --config config.yaml
```

Additional configurations are built for more complex models (AlexNet, ResNet, EfficientNet), but require a custom dataloader. Custom dataloders can be built by modifying the cnnMCSE repository. See the cnnMCSE repository for more details.

## Materials

*Algorithm Implementation for Medical Imaging*

We have utilized three large public chest radiograph datasets in this study: MIMIC-CXR, CheXpert, and ChestX-ray14 ([Table 1](#)). The MIMIC-CXR dataset was collected from Beth Israel Deaconess Medical Center (Boston, MA, United States) between 2011 and 2016, the CheXpert dataset was collected from Stanford Hospital (Stanford, CA, United States) between October 2002 and July 2017, and the NIH ChestX-ray14 dataset was collected from the NIH Clinical Center (Bethesda, MD, United States) between 1992 and 2015. We focus on diagnostic labels with sufficiently high frequencies.

A standard practice in deep learning on medical imaging classification tasks is the use of transfer learning – utilization of a pre-trained model such as AlexNet which is then fine-tuned on another dataset for a given task. We employ transfer learning baseline architectures and fine-tune the classification block to generate our classifiers, and the encoder-decoder blocks to generate our autoencoders. We seed each experiment with PyTorch, bootstrap experiments at 50x and report the Area Under the Curve Metric on a held-out test dataset (AUC) for our classifiers and the $L_2$-norm for the autoencoders. Code is provided as an attachment.

To compute the AEq values, we utilized standard machine learning practices to generate an autoencoder with tunable weights and biases. We generated joint datasets by merging a balanced sample of chest X-rays belonging to each race, controlling for sampling bias across different groups. We adapted a pre-trained RadImageNet[5], which has 60 million parameters for the convolutional and max pooling layers. We replaced the linear layers with an autoencoder with a latent space h = 2, selected via a grid-search based method. The autoencoder loss was trained via mean-squared error or $L_2$-norm. The network was trained via Adam optimization, with a learning rate determined by a simple grid search. All experiments were bootstrapped fifty times, and error bars represent 95% Confidence Intervals. The data-loaders and networks classes were generated with PyTorch. All experiments were trained on a single NVIDIA A100 GPU using the CUDA toolkit backend. All values and 95% confidence intervals are reported. P values are unpaired t-tests for two groups, and ANOVA for multiple groups.

We used a pretrained RadImageNet model to generate encodings for the Chest X-ray. We added three dense layers and a SoftMax layer to the top of the model to generate a single-class classifier. We used a simple grid search with values ranging from $10^{-5}$ to $10^{-1}$ for the learning rate and 4 to 256 for the batch size. After the hyperparameter optimization, we fine-tuned the dense layers and classification layers to generate classifications on a training set that was guided by AEquity values in comparison to naïve dataset collection. For example, in the case of the pneumothorax, the AEquity value dictated a diversification of the dataset, which was a 50-50 split between black and white patients, whereas in the case of edema, the AEquity value required population prioritization, which meant collecting samples belonging to black patients exclusively.

The train-validation-test split was 60-20-20, with the size of the training dataset being 30,000 samples from the MIMIC Chest-X-ray dataset. We trained for a maximum of 30 epochs and employed early stopping on validation area-under-the-curve metrics. For each experiment, we bootstrap the train and validation datasets with different seeds, and show that our results for generalizability are robust to sampling. The train, validation and test set, which consist of mutually exclusive samples, are bootstrapped fifty times and 95% confidence intervals are reported.

We take a data-centric approach with simple training of a ResNet-50 pretrained model and achieve similar performance to state-of-the-art approaches with use self-supervision and knowledge distillation, label dependencies, hand-crafted features, location aware dense networks. We report the results from the knowledge distillation paper below, which was benchmarked against the latter methods (**Supplementary Table S2**).

We define bias as the discrepancy between generalization performance (as measured by the area under the receiver operating characteristic curve) for two groups, for example, White and Black patients. We choose this over traditional fairness metrics like precision or false negative rates because AEquity works directly with the dataset and doesn't concern itself with model calibration. We later show empirically that bias mitigation with AEquity is robust to the choice of fairness metric (i.e., bias as measured by traditional fairness metrics is mitigated as well).

We calculate % reduction in bias with the following steps.
First, we calculate the bias without a data-centric intervention.

$$Bias_{pre} = (Test\ AUC_{pre,white} - Test\ AUC_{pre,black})$$

Second, we calculate the bias with a data-centric intervention.

$$Bias_{post} = (Test\ AUC_{post,white} - Test\ AUC_{post,black})$$

Third, we calculate the reduction in bias as a percentage of the original bias.

$$Bias\ Reduction = 100 \times \frac{Bias_{Pre} - Bias_{post}}{Bias_{Pre}}$$

We report the reduction in bias for both diversification and population prioritization.

## Algorithm Implementation for Cost-Utilization Analysis

In the data provided by (4), the data frame has two elements.
1. $Y_{it}$: Outcome – Avoidable Cost, Total Cost or Comorbidity Score.
2. $X_{i,t-1}$: The data collected by the patients insurer over the year *t-1*.
   a. Demographics (age, sex excluding race)
   b. Insurance type

  c. ICD-9 diagnosis and procedure codes

  d. Prescribed medications.

  e. Encounters

  f. Billed Amounts.

  From these, 149 features which include biomarkers, demographics and individual datasets are generated. The output feature is either the total cost, avoidable costs due to emergency visits and hospitalizations, and health which measures how many chronic conditions are flaring up and driving utilization. Those whose metrics exceed a lower threshold, the 50th percentile, are typically referred to their primary care physicians, who are provided with additional metrics about the patients and prompted to consider whether they would benefit from enrollment. In our algorithm, we assign them to the high-risk category.

  To compute the AEquity values, we utilized standard machine learning practices to generate an autoencoder with tunable weights and biases. We generated combined datasets by merging a balanced sample of electronic health record-derived data belonging to each race, controlling for sampling bias across different groups. The tabular nature of the dataset requires a fully connected network encoder and decoder, which was trained via Adam optimization, with a learning rate determined by a simple grid search. All experiments were bootstrapped fifty times with different seeds that initialized different weights with Xavier initialization. The error bars represent 95% confidence intervals. All experiments were trained on a single NVIDIA A100 GPU using the CUDA toolkit backend. All values and 95% confidence intervals are reported. P values are unpaired t-tests for two groups, and ANOVA for multiple groups.

  We provide a notebook demonstration and YouTube tutorial on how to use this method for any tabularized dataset, using the EHR dataset as the example[6]. Selecting an informative label on the electronic health record dataset can improve test AUC by 0.11 ($\Delta$AUC = 0.109; 95% CI: (0.106 – 0.112); $P < 0.05$)) on Black Patients by decreasing false negative rate and increasing precision.

# Supplementary Text

## Replication on Age and Sex

In this section, we demonstrate the potential of using the AEquity metric to understand biases in age and sex, which are demographic variables provided in all three datasets containing chest radiographs. For the MIMIC-CXR dataset, we notice that while most diagnoses (5/9) have a higher AEquity value for females over males, the labels that appear at higher frequencies such as opacity and effusion tend to have a higher AEquity value for males rather than females. However the joint AEquity values for these labels – opacity and effusion – tend to be lower than the AEq on individual groups, which may mitigate the biases of these labels in practice. In the CheXPert dataset, we see that only three labels appear to have statistically significant differences in AEquity values between males and females – cardiomegaly, enlarged cardio-mediastinum, and consolidation. Two of these labels have higher AEquity values for females (cardiomegaly and enlarged cardio-mediastinum), and one has a higher AEquity value for males (consolidation). In the NIH dataset, we see that while females have a lower AEquity value across the in five out of the seven diagnoses, the joint AEquity values are significantly larger in four of these, which means that training on a joint dataset may perpetuate some of the biases, and it makes more sense to further prioritize collection of data on female individuals.

Next, we look at the effect of age on AEquity values. In the MIMIC-CXR, we see that an older age corresponds to a higher AEquity value in three of the diagnoses – enlarged cardiomediastinum, pneumonia and pneumothorax, and a younger age corresponds to a higher AEquity value for edema and cardiomegaly. In the ChestX-ray14, we see that generally AEquity is generally balanced with a higher AEquity value for younger individuals in two out of the seven labels (opacity and pneumothorax). In the CheXPert data, we see that age is relatively balanced with a higher AEquity value in two out of the nine diagnoses for the younger age – pneumonia and effusion.

## Investigations and Interventions based on Medicare Status

We sought to validate our interventions in comparisons of insurance status. First, however, we investigated the coherence between age-related biases, as measured by Medicare status, and the race-related biases that we previously measured. We wanted to see if the impacts of age and race on biases, measured by AEquity, are coherent or dis-coherent.

We draw data from individuals with Medicare and those individuals with private insurance in Whites only, to control for the effects of race on data distribution observed above. We see some differences in distribution between the privately and Medicare-insured individuals. For example, chest radiographs labeled with pneumothorax in individuals with Medicare have a significantly higher AEquity value than those in individuals that are privately insured ($AEq_{Medicare, Pneumothorax}$ = 9.89; 95% CI: (9.80, 9.90), $AEquity_{Private, Pneumothorax}$ = 8.71; 95% CI: (8.64, 8.78); *P < 0.01*) (**Fig S3(a), 3(b)**). Moreover, the joint AEquity value is strictly in between the two different insurance status, which may suggest that there is some but not all overlap between the two distributions ($AEquity_{Joint, Pneumothorax}$ = 8.95; 95% CI: (8.87, 9.04)). This difference was observed in ethnicity bias experiments, which may suggest that the differences in distribution are somewhat captured by inequities in care that are overlapping between Black individuals and White individuals that are insured on Medicare. However, the joint AEquity values of the datasets merged by ethnicity were slightly lower than either of the individual ethnic group's AEquity values, which indicates that Blacks and Whites had overlapping distributions ($AEquity_{Joint, Pneumothorax}$ = 9.44; 95% CI: (9.35, 9.52), $AEquity_{Black, Pneumothorax}$ = 10.1; *95% CI*: (10.01, 10.18), $AEquity_{White, Pneumothorax}$ = 9.50; *95% CI*: (9.41, 9.57), *P > 0.05*). When stratifying by insurance status, the joint AEquity is between the two individual insurance status, which indicates that there is a significant complexity bias in the distribution of pneumothorax when stratifying by privately and Medicare-insured individuals.

We see the opposite trend in chest radiographs labeled with a diagnosis of atelectasis where privately insured individuals have a significantly larger AEquity value than individuals on Medicare ($AEquity_{Private, Atelectasis}$ = 8.82; *95% CI:* (8.72, 8.90), $AEquity_{Medicare, Atelectasis}$ = 8.08; 95% CI: (8.00, 8.15); *P < 0.05*). The joint AEquity value is between the two distributions, which indicates a non-overlapping data distribution ($AEquity_{Joint, Atelectasis}$ = 8.59; 95% CI: (8.50, 8.67)). In contrast, Black individuals have an AEquity value higher than White individuals for this specific diagnosis, which indicates that this trend cannot be solely explained by systemic disparities in healthcare ($AEquity_{Black, Atelectasis}$ = 9.12; 95% CI: (9.01, 9.22), $AEquity_{White, Atelectasis}$ = 8.22; 95% CI: (8.13, 8.30), *P < 0.05*).

Next, we highlighted that our interventions, which initially were conducted on race, are valid in age as measured by Insurance status (**Fig S3(c), S3(d)**). or effusion, $AEquity_{joint}$ is smaller than $AEquity_{private}$, $AEquity_{Medicare}$. ($AEquity_{Private, Effusion}$ = 8.08; 95% CI : (8.00, 8.16), $AEquity_{Medicare, Effusion}$ = 7.81; 95% CI: (7.73, 7.89), $AEquity_{Joint, Effusion}$ = 7.65; 95% CI: (7.56, 7.73)). In the case where the AEquity value requires us to increase dataset diversity, we add samples to the training dataset from each group equally – drawing equally from patients with and without private insurance. In the case where the AEquity value recommends population prioritization, we add samples exclusively from patients not on private insurance. We compare both to the naïve approach of data collection. For effusion at the largest evaluated sample size, there was a 0.02 increase in test AUC on chest X-rays belonging to individuals with Medicare following the intervention, ($AUC_{pre}$ = 0.79; *95% CI*: (0.79, 0.80) vs. $AUC_{post}$ = 0.83; *95% CI*: (0.82, 0.84) n = 8,192). For pneumothorax, $AEq_{joint}$ is larger than $AEquity_{private}$ but smaller than $AEquity_{Medicare}$ ($AEquity_{Joint, Pneumothorax}$ = 8.96; 95% CI: (8.87, 9.04), $AEquity_{Private, Pneumothorax}$ = 8.71; 95% CI: (8.64, 8.78), $AEquity_{Medicare, Pneumothorax}$ = 9.89; 95% CI: (9.80, 9.99), P < 0.05). Population prioritization leads to a 0.06 increase in test AUC for pneumothorax on Chest X-rays belonging to individuals with Medicare (AUC$_{Pre}$ = 0.59; 95% CI: (0.59, 0.60) vs AUC$_{Post}$ = 0.65; 95% CI: (0.65, 0.66), P < 0.05 n = 8,192)

## Intersectional Analysis on Black Patients and Age, Socioeconomic Status, and Sex

In this analysis, we investigate the intersection of race and age, socioeconomic status and sex (**Fig S4**). We run AEquity analysis on datasets that intersect between different under-represented groups including age, socioeconomic status and sex. First, we notice that for age, we compared patients that were between the ages of 20-40 and the ages of 40-60. For the analysis on the intersection of age and race, we see that AEq value is significantly higher for the older group in 5 out the 9 positive labels, versus higher for the younger group in 2 out of the 9 positive labels. The "No finding" label is higher for patients in the younger demographic, which may be explained by significant more variability in patient presentation where X-ray may have been ordered as a rule-out diagnosis rather than to specifically attribute a diagnostic cause. However, we notice that the joint is lower than that for older individuals for the labels Opacity, None, Pneumothorax, Edema and Atelectasis, which may mean that training on a joint dataset may mask some of potential biases we could have seen between the groups.

Second, we investigate the role of the intersection between sex and race. We analyze the chest X-rays in black male patients and black female patients. We previously saw that white female patients generally have a higher AEquity value than their male counterparts in the MIMIC Chest X-ray dataset This trend also holds true in black patients. We see that in four of the 9 labels, AEquity values are significantly higher for females than males, whereas we see that black male patients have a significantly higher AEquity in three of the 9 labels. However, in these three labels, we see that the joint AEquity is larger than AEquity values of the individual subgroups (Opacity, Cardiomegaly and Effusion), which means that these populations have different distributions for these labels. Training on a balanced dataset or one that may have been sampled more black male patients, for example, would have still induced generalizability biases on the under-represented population.

In our third analysis, we looked at the intersection between black patients on private insurance and black patients on Medicaid. Here, we notice the biggest potential biases occur when socioeconomic status is stacked on race. Looking simply at 'no diagnosis', we see that the AEquity for black patients on Medicaid is 10.8, whereas the AEquity for black patients that are privately insured is 9.2. This difference is both statistically significant (P < 0.05) and clinically significant because it implies there is a significant amount of complexity that occurs at baseline when comparing black patients on Medicare to black patients on private insurance.  Second, we notice that the AEquity values are significantly higher for black patients on Medicaid than black patients that are privately insured in five of the class labels, but the opposite is only true for a single label, while the remaining labels are neutral. The intersection between race and socioeconomic status potentiates the largest potential for generalizability error, when compared to the intersectionality of race and sex or race and age.

Finally, we apply the relevant dataset intervention to Black patients on Medicaid with respect to the 'No Finding' or 'None' diagnosis. The AUC increases significantly, and we see that this change in the AUC increases by approximately 0.03 and is statistically significant ($P < 1 \times 10^{-5}$). To further dissect where this increase in AUC is coming from, we sought to further investigate the different fairness metrics including True Negative Rate, True Positive Rate, False Positive Rate, as well as further derived fairness metrics such as Precision, False Discovery Rate, and Predicted Prevalence. We see that there is a clinically and statistically significant decrease in false positive rate for the diagnosis of "No Finding" ($\Delta FPR = 0.10$) and a corresponding increase in true negative rate ($\Delta TNR = 0.10$), with a clinically insignificant difference in false negative rate ($\Delta FNR = -0.01$) and true positive rate ($\Delta TPR = -0.01$). A lower false positive rate for the diagnosis of "No Finding" is equivalent to having a lower under-diagnosis bias rate. Furthermore, this translates into a higher precision, lower false discovery rate, and lower predicted prevalence. Changing the dataset gets at the root of the model's problem described previously, which was under-diagnosis of patients in under-represented models. By taking the relevant mitigation measures, AEquity can reduce the false discovery rate and predicted prevalence of "No Finding", thereby reducing the under-diagnosis bias in intersectional under-represented minorities.

## Analysis for Asian patients

In the MIMIC-CXR dataset, there are multiple races represented at lower sampling frequencies than White including Black and Asian. However, models trained on chest X-rays perform worse on Black patients compared to Asian patients. According to an analysis of AEquity values, this is because the sampling bias is compounded by complexity bias in the case of black patients, whereas in Asian patients, the complexity bias may be less severe if present at all. When we calculate AEquity values with respect to Asian patients, we notice two things – first that the AEquity values for Asian patients tend to be lower than those of White patients. Second, the joint AEquity values between the Asian patients and White patients are also lower, indicating sampling bias. This is because the chest X-rays of Asian and White patients are also relatively similar compared to the difference between X-rays for Black and White patients. We have computed the AEquity for these patients below and reported them in Supplemental Figure 5.

## Supplementary Figures

Figure S1. Replication on sex demographic

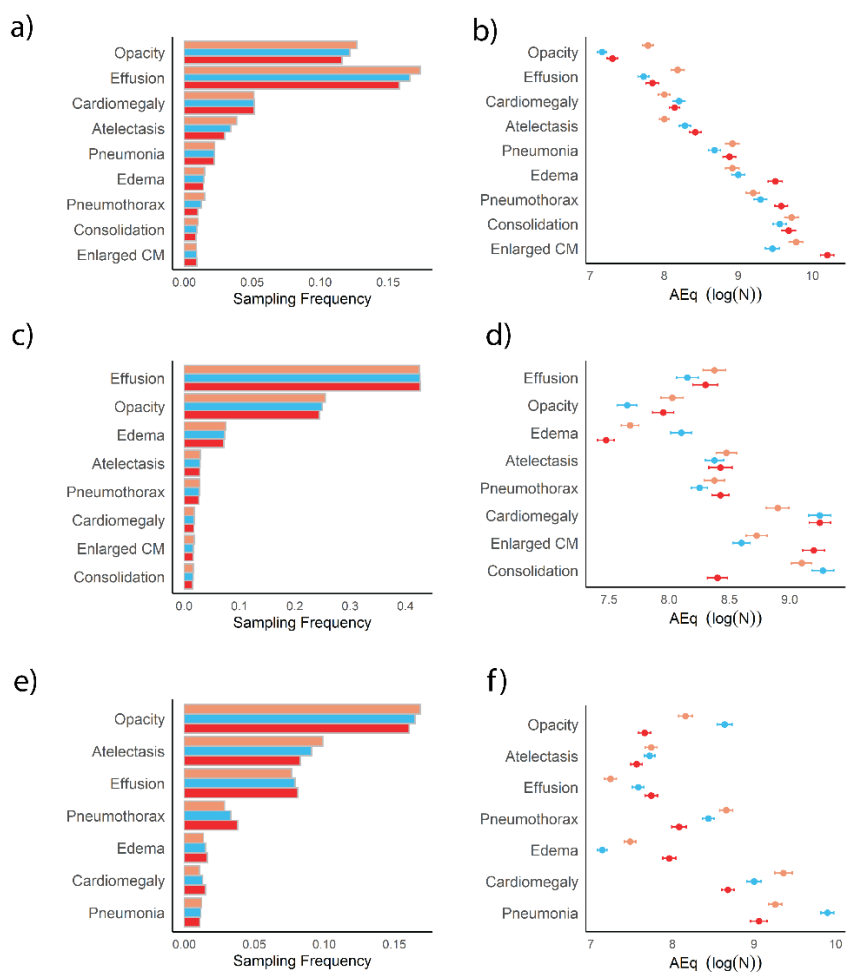

**Fig. S1**. **Replication on sex demographic**. Red is female, orange is male and blue is the combined dataset. **a)** Histogram by class label of MIMIC-CXR and **b)** AEquity values for MIMIC-CXR. (R = -0.80, P = 4.94 x $10^{-7}$) **c)** Histogram by class label for CheXPert and **d)** AEquity values by class label for CheXPert. (R = -0.41, P = 8 x $10^{-3}$). **e)** Histogram by class label for ChestX-ray14, and **f)** Corresponding AEquity values for ChestX-ray14 (R = -0.41, P = 7 x $10^{-3}$).

Figure S2. Replication on Age Demographic

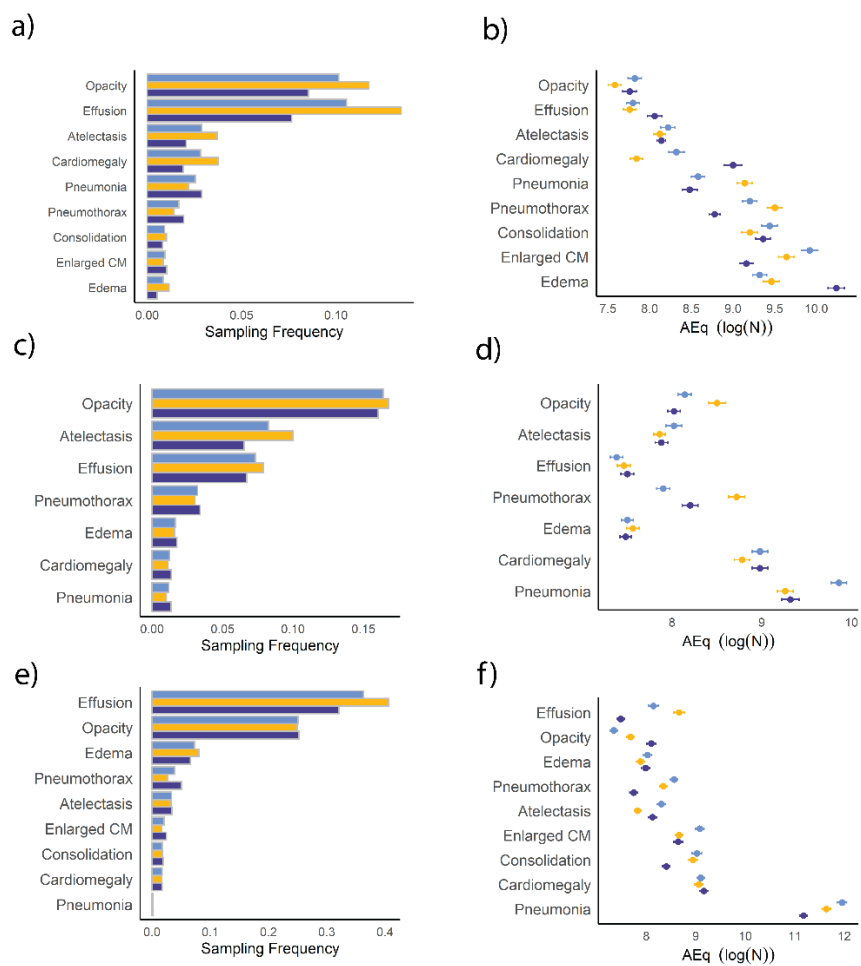

**Fig S2. Replication on age demographic.** Dark blue consists of individuals age (20-40) and yellow consists of individuals from 40-60. Light blue represents the joint dataset **a)** Histogram for individuals by age on MIMIC-CXR, **b)** Corresponding AEquity values for age on MIMIC-CXR (R =-0.88, P = 2 x $10^{-8}$). **c)** Histogram for individuals by age for ChestX-ray14 and **d)** corresponding AEquity values (R=-0.42, P = 3.4 x $10^{-5}$ ). **e)** Histogram for individuals by age on CheXPert and **f)** corresponding AEquity values (R = -0.32, P = 9.2 x $10^{-3}$).

Figure S3. AEquity applied to the MIMIC-CXR dataset stratified by Insurance Status.

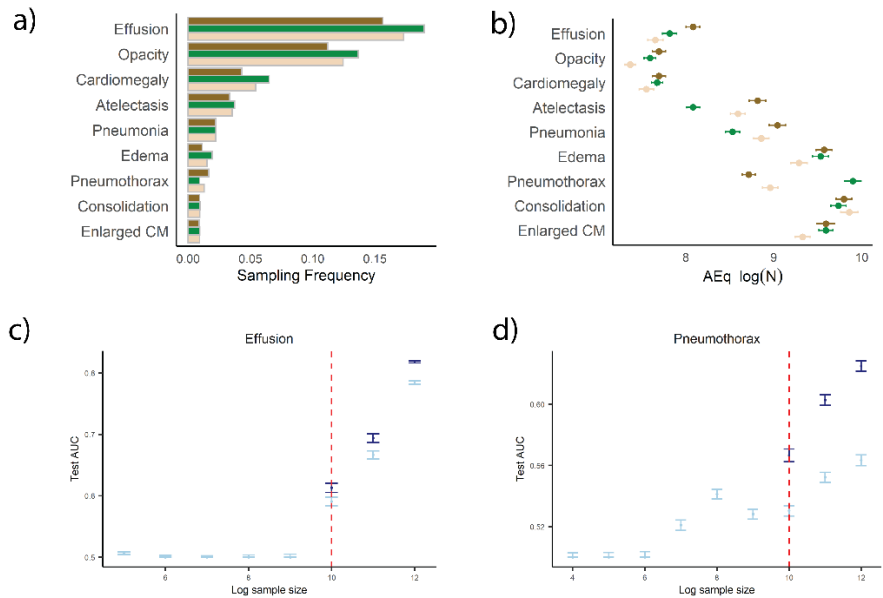

**Fig S3**. **AEquity applied to the MIMIC-CXR dataset stratified by Insurance Status.** Dark green represents privately insured, brown represents insured by Medicare, and beige represents joint data. **a)** Histogram of diagnostic frequencies by Insurance Status **b)** AEquity values by Insurance status. **c)** Intervention to increase dataset diversity improves performance in individuals on Medicare with effusion. **d)** Intervention to prioritize populations improves performance significantly in individuals on Medicare.

Figure S4. AEquity Applied to Intersectional Populations.

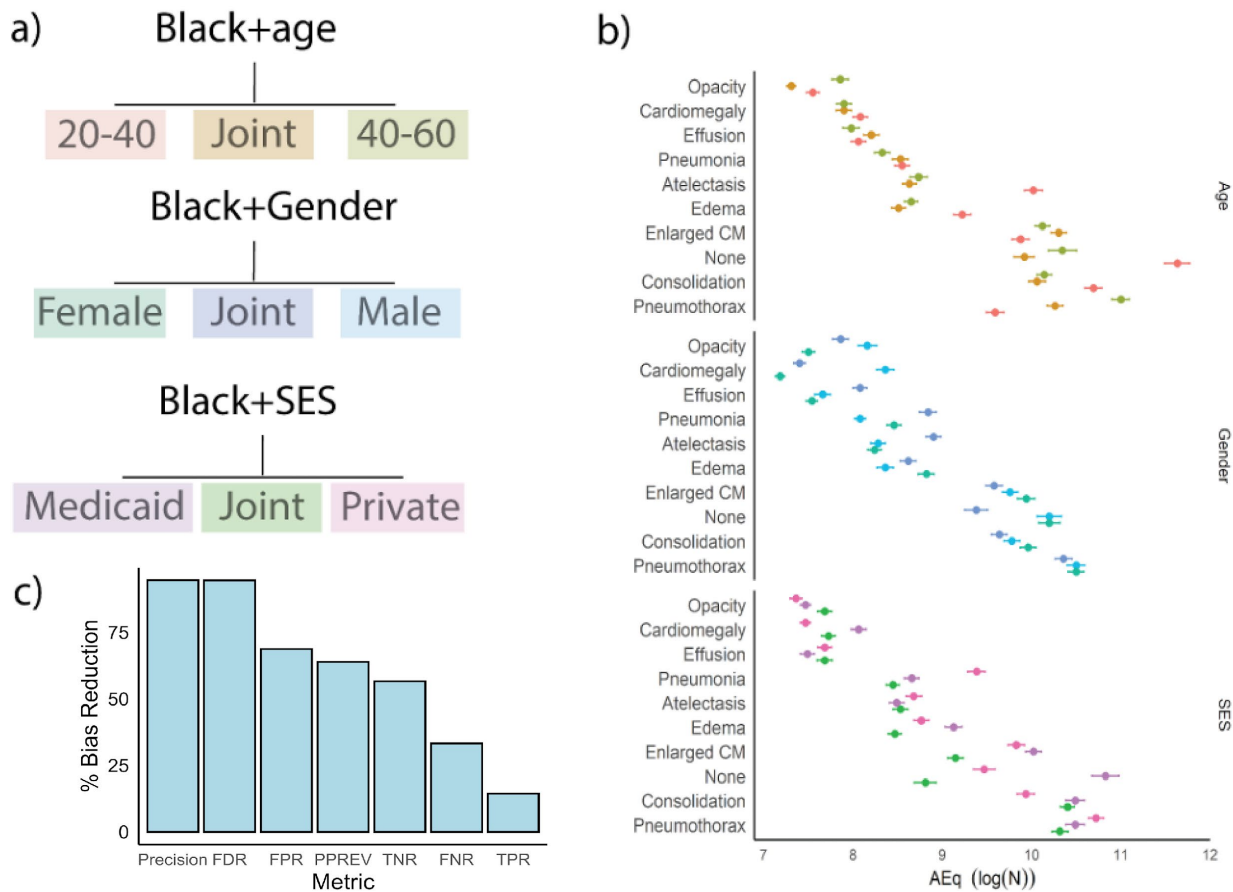

**Fig S4. AEquity applied to intersectional patient populations.** (a) Age – Black patients aged 20-40 are colored red, Black patients aged 40-60 are colored green, and the combined dataset is colored gold. Sex – Male black patients are colored lighter turquoise, Female black patients are colored light green, and the combined dataset is colored dark blue. Socioeconomic Status – Black patients on private insurance are colored pink, Black patients on Medicaid are colored dark purple, and the combined dataset is colored dark green. (b) AEq values calculated for each group across each label. (c) Interventions applied to black patients on Medicaid and the resulting effects on different fairness metrics.

Figure S5. AEquity applied to patient population identifying as Asian.

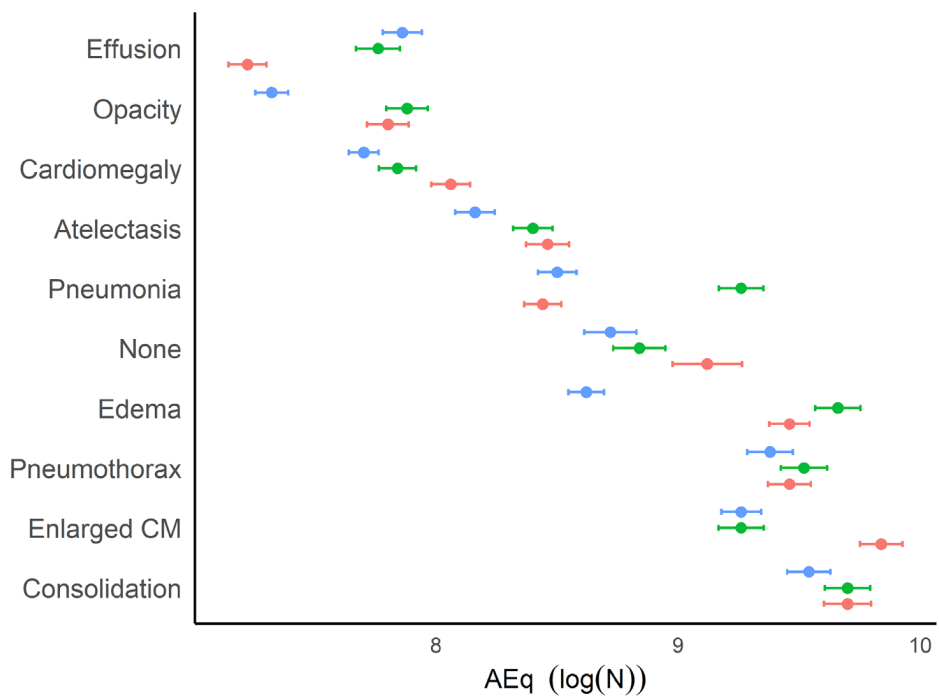

**Fig S5. AEquity applied to patient population identifying as Asian.** The bright red represents AEquity values calculated for individuals identifying as Asian. The green represents AEquity values calculated for individuals identifying as White, and the blue represents AEquity values calculated for a combined subset of individuals identifying as either Asian or White individuals. The AEquity of the combined subset is smaller than the individual subsets in eight of the 10 labels.

# Supplementary Tables

Table S1. Description of Datasets Utilized in Paper.

| Dataset | Description | Label/Outcome | Groups examined |
|---|---|---|---|
| MIMIC-CXR | 377,110 chest radiographs of 65,379 patients from the Beth Israel Deaconess Medical Center between 2011 and 2016. | Opacity, cardiomegaly, effusion, pneumonia, atelectasis, edema, enlarged cardiomediastinum, consolidation, pneumothorax, no finding. | Race, sex, age, insurance status; Sex, age and insurance in Black patients. |
| Healthcare costs | All primary care patients (N = 49,618) enrolled in risk-based contracts from 2013 to 2015, from an unspecified large academic hospital. Data included demographics, visit information, costs, comorbidities and biomarkers. | Total costs, avoidable costs, comorbidity score. | Race. |
| ChestX-ray14 | 108,948 frontal chest radiographs from 32,717 patients from the NIH Clinical Center between 1992 and 2015. | Atelectasis, cardiomegaly, effusion, infiltration, mass, nodule, pneumonia, pneumothroax. | Sex, age. |
| CheXpert | 224,316 chest radiographs of 65,240 patients performed between October 2002 and July 2017 in both inpatient and outpatient centers affiliated with Stanford Hospital. | Opacity, atelectasis, effusion, pneumothorax, edema, cardiomegaly, pneumonia. | Sex, age. |

**Description of datasets examined in this paper.**

Table S2. Comparing AEquity to SOTA methods on MIMIC Chest X-ray classification tasks.

| Method | Test Area-Under-Curve |
|---|---|
| RadImageNet (2020)[5] | 0.71 |
| Wang et al. (2020)[7] | 0.74 |
| Ho et al. (2020)[8] | 0.74 |
| Yao et al (2018)[9] | 0.78 |
| Ours | 0.74 |

**Comparing AEquity to SOTA methods on MIMIC Chest X-ray classification tasks.** We take a data-centric approach with simple training of a ResNet-50 pretrained model and achieve similar performance to state-of-the-art approaches that utilize a ResNet-50 backbone.

Table S3. Measurement of bias and intervention effect by metric on an Intersectional Population

| Metric | Bias Reduction (%) | Error (SE) |
|---|---|---|
| True Positive Rate | 14.4 | 0.04 |
| True Negative Rate | 56.6 | 2.35 |
| False Positive Rate | 68.7 | 0.14 |
| False Negative Rate | 33.3 | 3.37 |
| Precision | 94.6 | 0.04 |
| False Discovery Rate | 94.5 | 0.09 |

We show the effect of the interventions on various metrics used to measure fairness for a patient population that is both Black and on Medicaid compared to White patients. With a dataset intervention, we have reduced the discrepancy between the two groups.

Table S4. AEquity metric by race and predicted risk level, for each outcome

| Metric | Category | Mean AEquity in Black (SD) | Mean AEquity in White (SD) | Difference in AEquity (95% CI) |
|---|---|---|---|---|
| Total Costs | high risk | 7.13 (0.04) | 7.59 (0.06) | 0.47 (0.36-0.57) |
| Total Costs | low risk | 7.03 (0.03) | 7.84 (0.06) | 0.81 (0.73-0.90) |
| Avoidable Costs | high risk | 7.09 (0.04) | 7.50 (0.04) | 0.41 (0.32-0.49) |

| Avoidable Costs | low risk | 7.91 (0.05) | 7.09 (0.05) | 0.81 (0.71-0.92) |
| Active chronic conditions | high risk | 7.22 (0.05) | 7.28 (0.05) | 0.06 (-0.04 to 0.16) |
| Active chronic conditions | low risk | 6.97 (0.04) | 7.31 (0.04) | 0.34 (0.26-0.43) |

**AEquity metric by race and predicted risk level, for each outcome**. We show the calculated AEquity metric for highest risk and lower risk individuals for Black and White races, for each of the three outcomes (total costs, avoidable costs and active chronic conditions).

**Supplement References**